\documentclass[lettersize,journal]{IEEEtran}
\usepackage{amsmath,amsfonts}
\usepackage{algorithmic}
\usepackage{algorithm}
\usepackage{array}
\usepackage[caption=false,font=normalsize,labelfont=sf,textfont=sf]{subfig}
\usepackage{textcomp}
\usepackage{stfloats}
\usepackage[hidelinks]{hyperref}
\usepackage{url}
\usepackage{verbatim}
\usepackage{graphicx}
\usepackage{cite}
\usepackage{multirow}
\usepackage{threeparttable}
\usepackage{booktabs}
\usepackage{subfig}
\usepackage{amsthm,amsmath}
\usepackage{todonotes}
\usepackage{mathrsfs}
\usepackage{algorithm}
\usepackage{algorithmic}
\newcommand{\tpm}[1]{{$\pm$#1}}

\newcommand{\etal}{\textit{et al.}}
\usepackage[normalem]{ulem}

\begin{document}

\title{Decoupling Long- and Short-Term Patterns in Spatiotemporal Inference}

\author{Junfeng Hu, Yuxuan Liang, Zhencheng Fan, Li Liu, Yifang Yin, Roger Zimmermann,~\IEEEmembership{Senior Member,~IEEE}
\thanks{J. Hu, R. Zimmermann are with the School of Computing, National University of Singapore, Singapore. (email: junfengh@u.nus.edu, rogerz@comp.nus.edu)}
\thanks{Y. Liang is with INTR Thrust, Hong Kong University of Science and Technology (Guangzhou), Guangzhou, China. (email: yuxliang@outlook.com)}
\thanks{L. Liu is with the School of Big Data \& Software Engineering, Chognqing University, China. (email: dcsliuli@cqu.edu.cn)}
\thanks{Y. Yin is with Institute for Infocomm Research, A$*$STAR, Singapore. (email: yin\_yifang@i2r.a-star.edu.sg)}
\thanks{Z. Fan is with the Faculty of Engineering and Information Technology, University of Technology Sydney, Australia. (email: Zhencheng.Fan@student.uts.edu.au)}
\thanks{Y. Liang, L. Liu are the corresponding authors of this paper.}}



\maketitle

\begin{abstract}
Sensors are the key to environmental monitoring, which impart benefits to smart cities in many aspects, such as providing real-time air quality information to assist human decision-making. 
However, it is impractical to deploy massive sensors due to the expensive costs, resulting in sparse data collection.
Therefore, how to get fine-grained data measurement has long been a pressing issue.
In this paper, we aim to infer values at non-sensor locations based on observations from available sensors (termed spatiotemporal inference), where capturing spatiotemporal relationships among the data plays a critical role.
Our investigations reveal two significant insights that have not been explored by previous works. 
Firstly, data exhibits distinct patterns at both long- and short-term temporal scales, which should be analyzed separately.  
Secondly, short-term patterns contain more delicate relations including those across spatial and temporal dimensions simultaneously, while long-term patterns involve high-level temporal trends. 
Based on these observations, we propose to decouple the modeling of short-term and long-term patterns.
Specifically, we introduce a joint spatiotemporal graph attention network to learn the relations across space and time for short-term patterns. 
Furthermore, we propose a graph recurrent network with a time skip strategy to alleviate the gradient vanishing problem and model the long-term dependencies.
Experimental results on four public real-world datasets demonstrate that our method effectively captures both long- and short-term relations, achieving state-of-the-art performance against existing methods.
\end{abstract}


\section{Introduction}
\begin{figure}
  \centering
  \includegraphics[width=1.\linewidth]{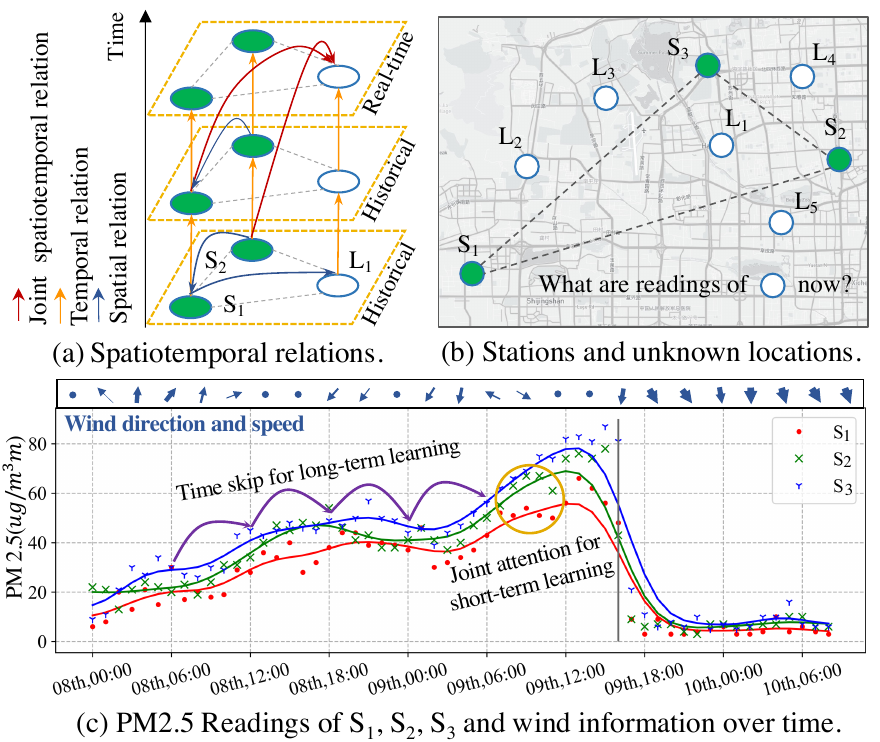}
  \caption{(a) Examples of three types of spatiotemporal dependencies. (b) Geographical distribution of stations and unknown locations. (c) PM2.5 readings and wind condition. The curves represent approximated trends and the arrows are time skips for long-term learning. The gray string denotes the time of 16:00, 09/11/2017 where signals do not tally with location relations.}
    \label{figure_intro}
\end{figure}

In recent years, numerous sensors have been deployed in different locations to sense the environment. They constantly report spatially-correlated and time-varying readings, such as traffic flows on roads and air quality measurements. Real-time monitoring of spatiotemporal data is of great importance to smart city efforts. For example, air quality information, e.g., the concentration of PM2.5 particles, can support air pollution control  and alert the public for health concerns. Unfortunately, one of the critical prerequisites for the above benefits is the fine-grained deployment of sensors, which usually leads to considerable expenditure and high energy consumption~\cite{zheng2013u}. Worse still, even existing sensors might lose readings due to factors such as a poor Internet connection. Thus, how to compensate for the pitfall of lacking sensors has become an urgent and challenging problem.
In this paper, we provide one solution by investigating the problem of spatiotemporal inference: given historical and real-time readings of existing sensors, we infer the real-time information \emph{at arbitrary locations} under a graph structure. As exemplified in Figure~\ref{figure_intro}(b), both historical and current readings of $S_1$--$S_3$ are leveraged to infer the real-time air quality status of locations $L_1$--$L_5$ without actual sensors in those places. 

 
Spatiotemporal inference requires delicate spatial and temporal dependency modeling~\cite{zheng2013u}. Early methods infer nodes based on linear dependencies, such as $k$-nearest neighbors (KNN) and inverse distance weighting (IDW)~\cite{lu2008adaptive}.
Then, non-linear relations are captured by subsequent approaches such as Gaussian processes (GPs)~\cite{rasmussen2003gaussian}.
However, the Gaussian assumption is rigid and the expensive computation also limits its applicability~\cite{appleby2020kriging}.
Recently, deep learning methods have emerged as a dominant paradigm. Among them, Spatio-Temporal Graph Neural Networks (STGNNs) are widely adopted due to their superior ability to handle non-Euclidean sensory data in graph structures~\cite{bruna2013spectral,hamilton2017inductive}.  
While the spatial relations among locations are naturally defined by the graph ~\cite{appleby2020kriging}, the temporal correlations among time points can also be captured, e.g., by concatenating a sequence of readings within a time window as the model input~\cite{wu2020inductive}.
In other related areas such as forecasting, approaches usually combine GNNs with recurrent neural networks (RNNs)~\cite{han2021joint} or temporal convolutional networks (TCNs)~\cite{wu2020connecting,zhao2019t} to learn spatial and temporal dynamics separately. While these methods effectively capture spatiotemporal dependencies, they encounter two major drawbacks when applied to the inference problem.

Firstly, they neglect the difference between long- and short-term patterns in the time series data. For example, readings can fluctuate within a short period in Figure~\ref{figure_intro}(c). On the contrary, when focusing on three curves from 00:00 AM to the next day at 12:00 PM, they still follow an increasing trend. This suggests that long- and short-term relations have inconsistent influences.
Unfortunately, the GNN-based inference method~\cite{wu2020inductive} concatenates temporal data as
features, ignoring this phenomenon. To resolve this, RNN might be feasible as the recurrent structure concentrates more on the latest frames. However, the gradient vanishing problem makes it hard to capture either long-term trends or delicate short-term patterns. Although different strategies, such as adopting adversarial training~\cite{yang2020adversarial}, are proposed to mitigate the problem, they are at the cost of computations and training difficulties.

Secondly, existing inference models lack the ability to effectively learn complex and dynamic spatiotemporal relationships. 
As illustrated in Figure~\ref{figure_intro}(a), readings are affected by both spatial relations in the graph (i.e., blue arrows) and its historical readings in the temporal dimension (i.e., orange arrows). 
Moreover, there exist more complicated \emph{joint spatiotemporal dependencies} (i.e., red arrows) that are influenced by sensors at different spatial and temporal positions directly. 
Unfortunately, the above GNN-RNN structures fail to explicitly consider them, which significantly hobbles the model's performance. Song~\etal~\cite{song2020spatial} attempt to capture the joint dependencies by a temporal-extended static graph structure. However, this static graph definition struggles to grasp the highly dynamic relations. To illustrate this, in Figure~\ref{figure_intro}(b) and (c), $S_2$ is geographically close to $S_3$ but around 16:00 of 09/11/2017, PM2.5 of $S_2$ is close to that of $S_1$, possibly due to the fickle wind condition. 
While several models are introduced to model dynamic relations~\cite{Shang2021discrete,qin2021dmgcrn,li2021spatial}, the majority of them focus on forecasting, and the inference problem is so far an under-explored research area.

To tackle these issues, we propose a \underline{Dual} Joint \underline{S}patio\underline{T}emporal \underline{N}etwork (DualSTN) for real-time spatiotemporal inference based on graph structures. 
Our DualSTN decouples short- and long-term learning into dual components: a \underline{J}oint \underline{S}patio\underline{T}emporal \underline{G}raph \underline{A}ttention ne\underline{T}work (JST-GAT) and a \underline{S}kip \underline{G}raph \underline{G}ated \underline{R}ecurrent \underline{U}nit (SG-GRU).
The first component adopts attention blocks to measure the impact between a node and \emph{its spatial neighboring nodes} within \emph{temporal short-term frames}, as the yellow circle shows in Figure~\ref{figure_intro}(c).
In this way, JST-GAT learns joint spatiotemporal relations explicitly, discarding the separate learning structures.
Meanwhile, impacts are measured by real-time sensor signals, which improves the method's ability to model potential dynamic relations. 
Inspired by Lai \etal~\cite{lai2018modeling}, the second component consists of a graph GRU with a time skip strategy, aiming to reach the same time span with fewer recurrent steps. This enables the model to capture the long-term temporal trends while ignoring dedicated short-term patterns, as the purple arrows illustrated in Figure~\ref{figure_intro}(c). 
For long-term dynamic relations, an intuitive way is to learn an adaptive adjacency matrix at each recurrent step as suggested by Wu \etal~\cite{wu2020connecting} and We \etal~\cite{wu2019graph}. However, their transductive design is incompatible with the inductive setting of our task where target locations are not involved during training. 
Thus, we improve the existing matrix learning method by making node embeddings rely on current input readings.
Additionally, we further leverage a graph sampling strategy to train the model~\cite{zeng2019graphsaint}, which further enhances its generalization ability.

We compare our model with state-of-the-art methods on four real-world datasets. Results show that our DualSTN outperforms the competitors clearly. To evaluate the effectiveness and influence of each module,
we also visualize the inference results and the attention weights to interpret DualSTN's ability on modeling long- and short-term patterns as well as dynamic spatiotemporal relations. Our code is available at \url{https://bit.ly/DualSTN} and the main contributions are summarized as follows:
\begin{itemize}
    \item We propose a new framework for spatiotemporal inference, which decouples the long- and short-term pattern learning into separate modules.
    \item We introduce a JST-GAT module that measures the interactions between nodes in different time and spatial dimensions concurrently, which captures the joint spatiotemporal relations explicitly.
    \item We propose an SG-GRU to facilitate long-term pattern modeling and optimization, where skip operations are introduced to maintain the same time span with fewer recurrent steps.
    \item Our DualSTN model achieves state-of-the-art performance on real-world datasets in diverse applications. These results demonstrate the effectiveness and generalization ability of our method.
\end{itemize}

\section{Related Work}

\subsection{Spatiotemporal Inference}
Spatiotemporal inference aims to infer signals of target locations with surrounding observed readings in a spatiotemporal domain. To solve the task, early statistical methods leverage linear relations modeling. For instance, $k$-nearest neighbors search and averages neighbor readings as the results while inverse distance weighting (IDW)~\cite{lu2008adaptive} further utilizes inverse distances as the weights. In addition, several approaches attempt to capture non-linear dependencies, and Kriging~\cite{rasmussen2003gaussian,cressie2015statistics} is one of the prevalent methods. Based on Gaussian Processes, it designs specialized kernels for the estimation of covariance between nodes and then infers targets by its posterior. However, the Gaussian assumption may not be followed by datasets and in this case, a transformation of non-Gaussian data is required~\cite{saito2000geostatistical}. Wallin \etal\cite{wallin2015geostatistical} attempt to map data into a geo-statistical configuration to weaken the assumption. 
Besides spatial relations, temporal dependencies are also taken into account. As an example, Yi \etal\cite{yi2016st} model them by hybrid variables derived from local and global spatiotemporal views. 
Alternatively, the problem can be regarded as anomaly detection~\cite{ozkan2015data} or matrix/tensor completion~\cite{bahadori2014fast}. Ozkan \etal~\cite{ozkan2015data} propose an anomaly detection algorithm that adopts a posteriori estimator to fill the missing data while
Yu~\etal~\cite{yu2016temporal} complete the matrix by a low-rank matrix assumption. 

Recently, deep learning methods have merged as a rife paradigm thanks to
their abilities on learning spatiotemporal relations in a data-driven way~\cite{smieja2018processing}.
Appleby \etal \cite{appleby2020kriging} propose a Kriging Convolutional Network (KCN) for spatial data inference, which adopts graph convolutional networks to extract dependencies from one-hop neighboring sensors. IGNNK~\cite{wu2020inductive} concatenates readings of a sequence along the channel dimension as the inputs of the model to further capture temporal relations. This design, however, treats temporal dependencies uniformly, ignoring inconsistency in the temporal relations. Recently, Wu \etal~\cite{wu2021spatial} propose SATCN consisting of a Spatial Aggregation Network with multiple aggregation functions to gather diverse spatial information. Then, TCNs are used to capture temporal dependencies. 
Some solutions concentrate on specific applications. For instance, Cheng \etal~\cite{cheng2018neural} describe a neural attention model using external features such as weather and POI, named ADAIN. 
Han \etal~\cite{Han0C21} introduce a novel multi-channel attention model (MCAM) that views external information as feature channels and utilizes LSTM for temporal modeling. However, these deep learning models learn spatial and temporal dependencies separately and external information is not always available, which limits the models' applications.

\begin{table*}
\center
\caption{Major characteristics of models. \# means the number of.}
\label{table_comparision}
  \begin{threeparttable}
  \begin{tabular}[width=0.98\linewidth]{lcccccc}
    \toprule
      \multirow{2}*{Model} & \multirow{2}*{Year} & Temporal Relation & \multicolumn{2}{c}{Spatial Relation} & \multirow{2}*{Learning Approach}\\
      \cmidrule(r){3-3}  \cmidrule(r){4-5}
      & & Method & Method & \#-hop  &  \\
      \midrule
      KNN & None & None & Linear & 1-hop & None \\
      IDW~\cite{lu2008adaptive} & 2008 & None & Linear & 1-hop   & None \\
      OKriging~\cite{cressie2015statistics} & 2015 & None & Gaussian & 1-hop & None \\
      GLTL~\cite{bahadori2014fast} & 2014 & Low-Rank Assumption  & Low-Rank Assumption & 1-hop & Transductive\\
      KCN~\cite{appleby2020kriging}  & 2020 & None & GNN & 1-hop  & Inductive\\
      IGNNK~\cite{wu2020inductive} & 2021 & GNN  & GNN & n-hop  & Inductive \\
      SATCN~\cite{wu2021spatial} & 2021  & TCN & Graph Aggregation &  n-hop & Inductive \\
      DualSTN (ours) & New & Joint Attention, GRU & Joint Attention &  n-hop  & Inductive\\
  \bottomrule
\end{tabular}
\end{threeparttable}
\end{table*}

\subsection{Spatiotemporal Graph Neural Network}
Spatiotemporal graph neural networks (STGNNs) are popular for spatiotemporal data modeling nowadays, following mainly two categories. They either couple GNNs with RNNs \cite{pan2020spatio,pan2019urban,xu2019spatio,shu2020host} or TCNs \cite{liu2023adversarial,wu2020connecting,liang2021fine,li2022online}. 
In the first category, GNNs are employed for capturing spatial dependencies, while RNNs are used to model temporal dynamics. For example, Li \etal~\cite{li2017diffusion} first propose a diffusion convolution that learns the spatial relations through bidirectional random walks on a graph and then capture temporal relations by RNNs. More advanced RNN models such as LSTM and GRU are also utilized in~\cite{xu2019spatio,ma2015long}. Xu \etal~\cite{xu2019spatio} aggregate representations from node neighborhoods as the inputs of a graph GRU while Lai \etal~\cite{lai2018modeling} use LSTM for long- and short-term modeling, which has a similar motivation to us but only focus on temporal relations.
In the second category, TCNs are adopted to learn temporal relationships and enjoy faster running speed than RNNs. For example, Liu \etal~\cite{liu2023adversarial} use the structure to identify more critical data and model it by the proposed adversarial algorithm.
Wu \etal \cite{wu2020connecting} propose a dilated inception temporal convolution to discover relations with different temporal scales.

In addition, attention mechanisms can be utilized to enhance the performance of STGNNS~\cite{vaswani2017attention,velivckovic2017graph,guo2019attention,ma2015long}. 
Zheng \etal~\cite{zheng2020gman} propose a multi-attention network to model spatial and temporal relations independently by attention. Wang \etal~\cite{wang2021multi} propose a multi-hop graph attention to calculate weights of context information from multi-hop neighbors. 
Cai \etal~\cite{cai2020traffic} explore data periodicity by dividing data into segments. The extracted segments are then fed into the attention network to capture temporal dependencies.
Huang \etal~\cite{huang2020lsgcn} combine GNNs and attention networks as a spatial gated block and adopt gated linear units (GLU) for temporal dimension, which achieved compelling performances for both short- and long-term forecasting tasks.
These designs, however, fail to capture the joint spatiotemporal relations that we aim to address. 

To model hidden relations that the adjacency matrix cannot reflect, Li \etal~\cite{li2021dynamic} and Bai \etal~\cite{bai2020adaptive} propose graph generation methods to learn an adaptive adjacency matrix and STGNNs take these two matrices to learn spatial relations. 
Unfortunately, the static structure cannot capture dynamic relations and their transductive learning approach is not suitable for our task.
To solve the challenge, Shin \etal~\cite{shin2022pgcn} progressively optimize the learned graph for new nodes that are not involved in training, based on their available readings. However, as readings of target locations are completely missing in our problem, this approach is also not applicable.

\subsection{Comparison to Existing Approaches}
We compare our model with other inference methods to highlight the differences in this section. KNN, IDW~\cite{lu2008adaptive}, and OKriging~\cite{cressie2015statistics} are statistical methods, while our DualSTN is a data-driven method. Meanwhile, OKriging is a geo-location method only applied to geographic data. On the contrary, our method is suitable for various datasets. 
The transductive GLTL~\cite{bahadori2014fast} is based on the low-rank assumption, while our model does not require any explicit assumption and is inductive.
For the deep learning methods, KCN and KCN-SAGE~\cite{appleby2020kriging} are one-hop models. Instead, our approach is an n-hop method and also takes temporal dependencies into consideration. IGNNK~\cite{wu2020inductive} adopts GNNs but ignores different long- and short-term patterns. SATCN~\cite{wu2021spatial} utilizes TCNs to capture temporal relations while our method decouples long- short-term learning and could model spatiotemporal dependencies simultaneously. Table~\ref{table_comparision} summarizes the model characteristics.

\section{Preliminaries}
\subsection{Problem Formulation}
In this work, we focus on the real-time spatiotemporal data inference task under a graph structure (see Figure~\ref{figure_intro} (a), (b)). A graph is represented by $\mathcal{G}=(\mathcal{V}, \mathcal{E}, \mathbf{A})$, where $\mathcal{V}$ is the node set, $\mathcal{E}$ is the set of edges and $\mathbf{A}$ is the pre-defined adjacency matrix. Suppose we have $N_o$ stations with observed spatiotemporal signals, we denote sensor signals at time $t$ as $\mathbf{X}_t=[\mathbf{x}^1_t, .., \mathbf{x}^{N_o}_t]\in \mathbb{R}^{N_o\times D}$, where $D$ is the number of attributes in a node. The goal aims to use available station readings $[\mathbf{X}_{\tau-T+1}, \mathbf{X}_{\tau-T+2}, .., \mathbf{X}_\tau]$ of time window $T$ to infer signals $\mathbf{Y}_\tau$ of $N_u$ locations at time $\tau$ given their spatial relations in the graph $\mathcal{G}$,
\begin{equation}
    [\mathbf{X}_{\tau-T+1}, \mathbf{X}_{\tau-T+2}, .., \mathbf{X}_\tau, \mathcal{G}]\stackrel{f_\Lambda(\cdot)}{\longrightarrow}[\mathbf{{Y}_\tau}],
\label{eq_gcn}
\end{equation}
where $f_\Lambda(\cdot)$ is the learned mapping function with parameters $\Lambda$ and assume $N = N_o + N_u$. Note that at any time $\tau$, we only use \emph{historical} and \emph{current} station readings $\mathbf{X}_{\tau-T+1:\tau}$ to infer target locations $\mathbf{{Y}}_\tau$, which follows the definition of real-time inference as no future readings are available. Further, it is possible that several stations lose readings due to a bad Internet connection or some sensors may be removed or added to the graph. This requires our model to be inductive to various numbers of stations and target locations by design. 

\subsection{Graph Convolution Layer}
As an essential operation to learn interactions among nodes defined by a graph structure~\cite{kipf2016semi}, the graph convolution aggregates node features from its neighbors to learn spatial correlations. By stacking convolution layers, the model is capable of learning dependencies from multi-hop neighbors to improve the modeling ability.
From the spatial perspective, a graph convolutional layer is formulated as:

\begin{equation}
    \mathbf{Z} = \phi(\mathbf{P} \mathbf{X} \mathbf{W}),
\label{eq_adpgcn}
\end{equation}
where $\mathbf{P}=\mathbf{D}^{-1} (\mathbf{A} + \mathbf{I}) \in \mathbb{R}^{N\times N}$ denotes the normalized adjacency matrix with self-loops, $\mathbf{D}$ is the degree matrix, $\mathbf{X}\in \mathbb{R}^{N\times D}$ are the input readings, $\mathbf{W} \in\mathbb{R}^{D\times F}$ are learnable parameters, $\phi(\cdot)$ is an activation function. Li \etal~\cite{li2017diffusion} further introduce a diffusion convolution that propagates graph features with $K$ steps:
\begin{equation}
    \mathbf{Z} = \phi(\sum^{K}_{k=1}\mathbf{P}^k \mathbf{X} \mathbf{W}_k).
\end{equation}
To capture hidden graph structures that the pre-defined adjacency matrix cannot reflect, Wu \etal~\cite{wu2019graph} propose an adaptive adjacency matrix learning method for the graph convolution, which results in:
\begin{equation}
    \mathbf{Z} = \phi(\sum^{K}_{k=1}{\mathbf{P}}^k \mathbf{X} \mathbf{W}_k + \mathbf{\hat{A}}^k \mathbf{X} \mathbf{U}_k),
\end{equation}
where $\mathbf{\hat{A}}$ is the adaptive adjacency matrix learned by a network. We term these two graph convolutions as $\Theta_{\star \mathcal{G}}(\mathbf{X}, \mathbf{A})$ and $\Theta_{\star \mathcal{G}}(\mathbf{X}, \mathbf{A}, \mathbf{\hat{A}})$, where $\Theta$ are learnable parameters. The important notations in the paper are reported in Table~\ref{table_notation}.

\begin{table}[h]
\caption{Import notations for data inference.}
\label{table_notation}
  \scalebox{0.9}{
  \begin{threeparttable}
  \begin{tabular}[width=0.78\linewidth]{ll}
    \toprule
      Notation & Description \\
      \midrule
    $N$, $N_o$, $N_u$ & number of nodes / observed sensors / target locations\\
    $T$ & time length for inference \\
    $t_s,t_k$ & short-term time window, number of skip steps\\
    $\mathbf{A}$, $\mathbf{\hat{A}}$ & pre-defined / learned adaptive adjacency matrix\\
    $\mathbf{x}^i_t$, $\mathbf{X}_t$, $\mathcal{X}$ & readings of the $i$-th sensor / all sensors at time $t$ / all sensors\\
    $\mathbf{\hat{Y}}_\tau^s$, $\mathbf{\hat{Y}}_\tau^l$ & inferred signals of short- / long-term learning at target time $\tau$\\
    $\Theta_{\star \mathcal{G}}(\mathbf{X}, \mathbf{{A}})$ & graph convolution only with pre-defined adjacency matrix\\
    $\Theta_{\star \mathcal{G}}(\mathbf{X}, \mathbf{{A}}, \mathbf{\hat{A}})$ & that with learned adaptive adjacency matrix\\
  \bottomrule
\end{tabular}
\end{threeparttable}
}
\end{table}

\section{METHODOLOGY}
\begin{figure*}
  \centering
  \includegraphics[width=0.98\linewidth]{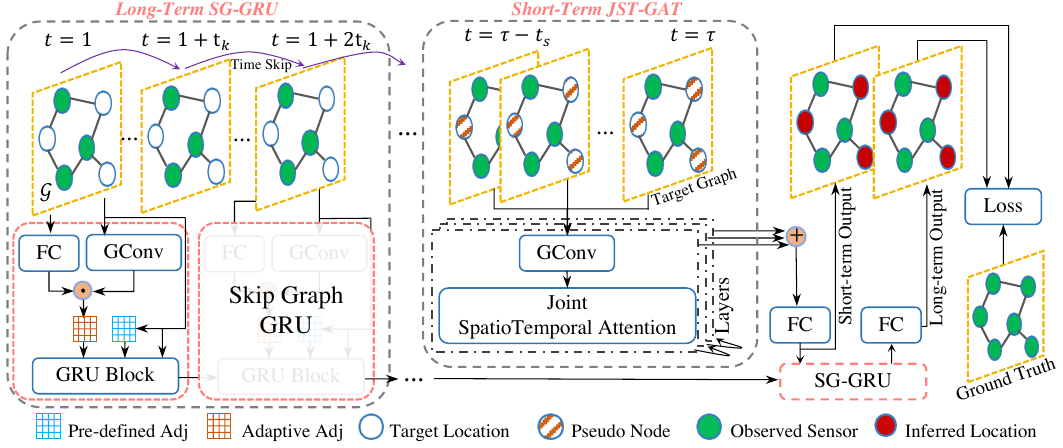}
  \caption{The framework of the proposed DualSTN model which contains two components: SG-GRU for long-term learning and JST-GAT for short-term learning. Adj means adjacency matrix. }
    \label{figure_framework}
\end{figure*}

\subsection{Overview of DualSTN}
Figure~\ref{figure_framework} illustrates the overall framework of our Dual SpatioTemporal Network (DualSTN) which consists of two backbone components for long- and short-term spatiotemporal patterns learning. The short-term Joint SpatioTemporal Graph Attention Network (JST-GAT) first generates \emph{pseudo nodes} for unknown locations for the following attention blocks. Then, stacked graph convolutions and spatiotemporal attention layers are used to learn joint spatiotemporal dependencies, followed by a fully-connected layer to generate short-term inference results. 
At each recurrent step, the long-term Skip Graph Gated Recurrent Unit (SG-GRU) first learns an adaptive adjacency matrix for the graph in an inductive approach. Then, the graph GRU further takes the learned matrix and hidden states to encode current readings. Last, it integrates short-term results as the input to generate long-term inference results. 
In the below sections, we first introduce the graph sampling strategy for inductive learning and then describe the details of DualSTN.

\subsection{Graph Sampling for Inductive Learning}
\begin{figure}[!h]
  \centering
  \includegraphics[width=1\linewidth]{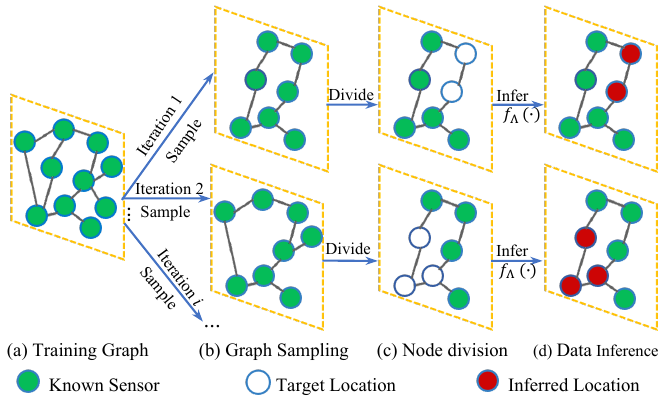}
  \caption{Illustrations of graph sampling. For each iteration, we sample a subgraph and divide nodes into observed sensors and unknown locations randomly.}
    \label{fig_sample}
\end{figure}

In a real-world environment, new sensors might be added to the graph and even existing sensors could retire after some time. In this situation, models need to be compatible with different graphs and input sensors. In addition, they ought to have a better generalization ability, which makes the task more challenging. Previous model Kriging Convolutional Network~\cite{appleby2020kriging} designed an inductive model but trained the model using a static graph, which is suboptimal and prone to overfitting. 
Nowadays, approaches solve this by either learning node embedding functions~\cite{hamilton2017inductive} or sampling subgraphs during training~\cite{zeng2019graphsaint,wu2020inductive}. In this paper, we adopt the sampling method which does not involve more parameters. As Figure~\ref{fig_sample} shows, given a training graph in (a), for each iteration, we first randomly sample a subgraph in (b). Then in (c), the subgraph is randomly divided into two groups dubbed observed sensors and target locations. In (d), we leverage observed sensors to infer signals of target locations and optimize the network. 
In this way, the model is less likely to be optimized according to knowledge from the absolute node locations and can capture universal spatiotemporal relations shared among sensors, strengthening its generalization ability.
Algorithm~\ref{algo_inductive} describes the graph sampling process in detail. After obtaining the subgraph, we can fetch a batch of data as the inputs for training. Note that we use $N$ to denote the number of nodes in the subgraph in the following sections.

\begin{algorithm}
\caption{Graph Sampling for Each Iteration}
\begin{algorithmic}[1]
\REQUIRE graph $\mathcal{G}=(\mathcal{V}, \mathcal{E}, \mathbf{A})$, sensor readings $\mathcal{X}$, where $\mathcal{X}\in \mathbb{R}^{T\times N\times D}$.
\ENSURE subgraph $\mathcal{G}_s$, sampled known sensor readings $\mathcal{X}_s$, sampled readings for inference $\mathcal{Y}_s$.
\STATE Randomly generate two integers $N_o$, $N_u$ indicating the number of known sensors and unknown locations such that $N_o + N_u \leq N$.
\STATE Sample a subset of nodes $\mathcal{V}_s$ from $\mathcal{V}$ such that $|\mathcal{V}_s|=N_o + N_u$
\STATE Divide $\mathcal{V}_s$ into two subsets $\mathcal{V}_k$ and $\mathcal{V}_u$ such that $|\mathcal{V}_k|=N_o$, $|\mathcal{V}_u|=N_u$ and $\mathcal{V}_k \cap \mathcal{V}_u = \emptyset$.
\STATE Retrieve sensor readings from two subsets: $\mathcal{V}_k \rightarrow \mathcal{X}_s$ and $\mathcal{V}_u \rightarrow \mathcal{Y}_s$.
\STATE Retrieve edges $\mathcal{E}_s$ from $ \mathcal{V}_s$.
\STATE Construct the adjacency matrix $\mathbf{A}_s$ using $\mathcal{E}_s$ and $\mathbf{A}$.
\RETURN $\mathcal{G}_s$, $\mathcal{X}_s$, $\mathcal{Y}_s$.
\end{algorithmic}
\label{algo_inductive}
\end{algorithm}

\subsection{Joint Spatiotemporal Graph Attention Network}
\subsubsection{K-nearest Inverse Distance Weighting}
\begin{figure}
  \centering
  \includegraphics[width=1\linewidth]{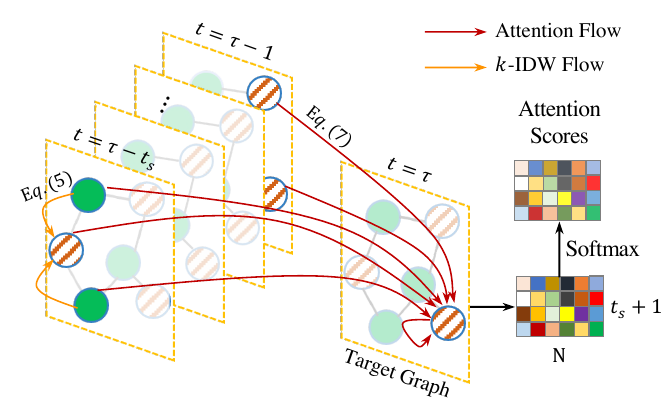}
  \caption{Illustrations and information flows of $k$-nearest inverse distance weighting and the joint spatiotemporal attention. Only a limited number of flows for one node are drawn.}
    \label{figure_sta}
\end{figure}

As we do not have signals of target locations in a graph, the attention mechanism is suboptimal to be applied directly. Thus, we first calculate initial readings for them by $k$-nearest inverse distance weighting ($k$-IDW). To be specific, we fill short-term values of the locations (termed \emph{pseudo nodes}) at time $\tau$ and its short-term neighbor frames in the window $[\tau-t_s, \tau-1]$. As shown in the yellow arrows of Figure~\ref{figure_sta}, $k$-IDW follows the idea of $k$-nearest neighbors that first searches the spatially $k$-nearest observed sensors for each target location. Then, the inverse distances from the location to its neighbors are utilized as weights to calculate the mean:
\begin{equation}
    \mathbf{\tilde{x}}_{t,i} = \frac{\sum\nolimits_{j=1}^{k} \mathbf{x}_{t,j}\odot {d_{i,j}}^{-\rho}}{\sum\nolimits_{j=1}^{k}{d_{i,j}}^{-\rho}},
\end{equation}
where $t\in [\tau-t_s, \tau]$, $k$ denotes the assigned number of nearest neighbors, $d_{i,j}$ is the distance between a tarrget location $i$ and the neighbor $j$, $\rho$ means the decay rate, and $\odot$ represents the Hadamard product. After obtaining pseudo nodes, they can be used to compute attention scores and we regard pseudo nodes and observed sensors uniformly, notated as $\mathbf{X}_t=[\mathbf{x}_{t,1}, .., \mathbf{x}_{t, N_o}, \mathbf{\tilde{x}}_{t,1}, .., \mathbf{\tilde{x}}_{t, N_u}] \in \mathbb{R}^{N\times D}$ below.

\subsubsection{Joint Spatiotemporal Attention}
Spatiotemporal data are influenced by conditions from surrounding locations which are highly interactive and difficult to capture the patterns. To learn these relationships, attention networks are proven to be an effective tool~\cite{vaswani2017attention}. Among them, researchers mainly utilize individual attention blocks to handle the spatial relations and temporal dynamics respectively, followed by a fusion module to integrate them~\cite{liang2018geoman,guo2019attention,zheng2020gman}. 
However, these models fail to explicitly consider joint spatiotemporal dependencies directly. For instance, assuming a north wind, PM2.5 particles will be blown toward the south over time. 
In addition, a car accident will clog traffic on an upstream road after a short time. The phenomena involve synchronous spatiotemporal shifts and are hard for separate attention modules to model.
We propose a joint spatiotemporal attention mechanism to capture them simultaneously, as illustrated in Figure~\ref{figure_sta}. 

Given features of a target frame $\mathbf{Z}^{l-1}_\tau$ at layer $l-1$ and its neighbor frames $\mathbf{Z}^{l-1}_t$ in the short-term window $[\tau-t_s, \tau-1]$, we first employ a graph convolution layer with skip connection to learn the spatial relations in each frame:
\begin{equation}
    \mathbf{Z}_t^l = \gamma \mathbf{Z}_t^{l-1} + \mu \Theta^l_{\star \mathcal{G}}\left(\mathbf{Z}^{l-1}_t, \mathbf{{A}}\right) \quad t\in [\tau-t_s, \tau],
\end{equation}
where $\mathbf{Z}^0_t=\mathbf{X}_t$, $\gamma$ and $\mu$ are hyperparameters for skip connection. Here, the adjacency matrix $\mathbf{{A}}$ defines static spatial relations as a prior so the attention mechanism relieves from learning it again, reducing learning difficulties. 
Then, the attention operation captures joint spatiotemporal dependencies. We first calculate attention weights between sensor $\mathbf{z}^l_{\tau, i}$ and other sensors $\mathbf{z}^l_{t,j}$ in the window formulated by:
\begin{equation}
    e^l_{t, i, j} = \mathbf{v_a^\top} \operatorname{tanh}(\mathbf{W_a}\mathbf{z}^l_{\tau,i} + \mathbf{U_a}\mathbf{z}^l_{t,j} + \mathbf{b_a}),
\end{equation}
\begin{equation}
    e^l_{t, i, j} = \frac{\exp(e^l_{t, i, j})}{\sum\nolimits_{t=\tau-t_s}^\tau\sum\nolimits_{j=1}^N\exp(e^l_{t, i, j})},
\end{equation}
where $\mathbf{v_a}, \mathbf{b_a}  \in\mathbb{R}^{F}$, $\mathbf{W_a}, \mathbf{U_a} \in \mathbb{R}^{D\times F}$ are learnable parameters. Note that the parameters are shared across all frames to reduce the computational cost.
We also compute attention scores within the target frame and in this situation, the block degrades into spatial attention. Finally, we obtain an attention map $\mathbf{E}^l\in \mathbb{R}^{(t_s+1)\times N\times N}$, where the second dimension $N$ refers to sensor features at the target time $\tau$ and the third dimension $N$ denotes sensors in the $t_s+1$ frames. 
Next, we use features of short-term frames and the attention map to learn short-term inference representations:
\begin{align}
\begin{split}
    &\mathbf{z}^l_{\tau,i} = \sum_{t=\tau-t_s}^{\tau} \sum_{j=1}^N \mathbf{E}^l_{t,i,j} \mathbf{z}^l_{t,j}.
\end{split}
\label{equation_gcn}
\end{align}

We stack the joint attention blocks for $L$ layers. On the top layer $L$, a fully connected layer is used to generate the short-term outputs: 
\begin{equation}
    \mathbf{\hat{Y}}_\tau^s = \mathbf{Z}^L_\tau \mathbf{W}_{fs} + \mathbf{b}_{fs},
\label{fl_out}
\end{equation}
where $\mathbf{W}_{fs}\in \mathbb{R}^{F\times D}$, $\mathbf{b}_{fs}\in \mathbb{R}^{D}$ are parameters and $\mathbf{\hat{Y}}_T^s\in\mathbb{R}^{N\times D}$ are short-term inference results.
Finally, the joint spatiotemporal dependencies can be learned by a single JST-GAT without separate modules. Moreover, the attention block aids interpretability by visualizing weights to understand how the model learns the spatiotemporal relations. 

\subsection{Skip Graph Gated Recurrent Unit}
\subsubsection{Inductive Dynamic Graph Generation}
To model the hidden relations among nodes, previous works learn an adaptive adjacency matrix during training and it remains static during testing~\cite{wu2019graph,bai2020adaptive,wu2020connecting}. However, this disregards the dynamic dependencies of the graph structure over the timeline. Later, Li \etal~\cite{li2021dynamic} model the dynamic connections by learning an adaptive matrix at each step of a recurrent network. Unfortunately, the method significantly relies on embeddings of training nodes which is not available in the inductive setting. To solve these challenges, we propose an inductive graph generation module that updates the adjacency matrix in an inductive fashion based on~\cite{wu2020connecting}. To be specific, at time step $t$, we use the historical hidden states $\mathbf{H}_{t-t_k}$, $\mathbf{X}_t$ and $\mathbf{A}$ to learn graph structure information. Here, the $t_k$ is a skip step described below. The node embedding is replaced by a fully-connected layer $FC(\cdot)$ taking $\mathbf{H}_{t-t_k}$ as input. In summary, the adaptive adjacency matrix $\mathbf{\hat{A}}_t$ at time $t$ is calculated by:
\begin{align}
\begin{split}
    &\mathbf{M}_t^{1}= \tanh(\Theta_{1\star \mathcal{G}}\left(\mathbf{X}_t, \mathbf{{A}}\right) \odot FC_1(\mathbf{H}_{t-t_k})),\\
    &\mathbf{M}_t^{2}= \tanh(\Theta_{2\star \mathcal{G}}\left(\mathbf{X}_t , \mathbf{{A}}\right) \odot FC_2(\mathbf{H}_{t-t_k})),\\
    &\mathbf{\hat{A}}_t = \operatorname{ReLU}(\tanh(\alpha(\mathbf{M}_t^{1}{\mathbf{M}_t^{2}}^\top - \mathbf{M}_t^{2}{\mathbf{M}_t^{1}}^\top))),
\end{split}
\end{align}
where $\mathbf{M}_1^{t}$ and $\mathbf{M}_2^{t} \in \mathbb{R}^{N\times F}$ are source node encoder and target node encoder, respectively, $\sigma(\cdot)$ is the sigmoid activation function and $\alpha$ is the saturation rate hyperparameter.

\subsubsection{Skip Graph Gated Recurrent Unit}
GRU is one of the recurrent structures designed to capture historical temporal information in a recurrent way~\cite{chung2014empirical}. However, the gradient vanishing and overwhelming state estimation of the latest inputs cause difficulties to capture long-term temporal patterns~\cite{bahdanau2014neural}. To alleviate this, motivated by~\cite{lai2018modeling} we propose a graph GRU with skips to maintain a temporal span with fewer recurrent steps. To be specific, hidden states are updated using historical hidden states of a certain number of skips $t_k$, which can be formulated as:
\begin{align}
\begin{split}
    &\mathbf{r}_t = \sigma(\Theta_{r\star \mathcal{G}}(\mathbf{X}_t | \mathbf{H}_{t-t_k}, \mathbf{A}, \mathbf{\hat{A}}) + \mathbf{b}_r),\\
    &\mathbf{u}_t = \sigma(\Theta_{u\star \mathcal{G}}(\mathbf{X}_t | \mathbf{H}_{t-t_k}, \mathbf{A}, \mathbf{\hat{A}}) + \mathbf{b}_u),\\
    &\mathbf{c}_t = \tanh(\Theta_{c\star \mathcal{G}}(\mathbf{X}_t | (\mathbf{H}_{t-t_k}\odot \mathbf{r}_t), \mathbf{\mathbf{A}}, \mathbf{\hat{A})} + \mathbf{b}_c),\\
    &\mathbf{H}_t = \mathbf{u}_t \odot \mathbf{H}_{t-t_k} + (1-\mathbf{u}_t)\odot \mathbf{c}_t,
\end{split}
\end{align}
where $|$ means concatenation, $\mathbf{r}_t$ and $\mathbf{u}_t \in\mathbb{R}^{N\times F}$ are reset gate and update gate, respectively. Through the recurrent skip, the module is encouraged to focus on the high-level long-term temporal patterns, which ignores delicate relations among consecutive frames. The decreased recurrent steps also facilitate the optimization process. At the last recurrent step at which the short-term inference results $\mathbf{\hat{Y}}_\tau^s$ are computed, we feed $\mathbf{\hat{Y}}_\tau^s$ to the graph GRU to compute the corresponding hidden states $\mathbf{H}_\tau$. Finally, a fully connected layer is used to obtain the long-term inference outputs:
\begin{equation}
    \mathbf{\hat{Y}}_\tau^l = \mathbf{H}_\tau \mathbf{W}_{fl} + \mathbf{b}_{fl},
\end{equation}
where $\mathbf{W}_{fl}\in \mathbb{R}^{F\times D}$, $\mathbf{b}_{fl}\in \mathbb{R}^{D}$ are parameters, and $\mathbf{\hat{Y}}_\tau^l\in\mathbb{R}^{N\times D}$ are long-term inference outputs. 

\subsection{Loss Function and Training Procedure}
To strengthen the generalization ability, we train our model by reconstructing all sensor signals instead of just target locations like Appleby \etal~\cite{appleby2020kriging} and simultaneously optimize the MAE loss of short-term outputs $\mathbf{\hat{Y}}_\tau^s$ as well as long-term results $\mathbf{\hat{Y}}_\tau^l$:
\begin{equation}
    \mathcal{L} = \frac{1}{N}\left|\mathbf{Y}_\tau-\mathbf{\hat{Y}}_\tau^l\right| + \frac{1}{N}\left|\mathbf{Y}_\tau-\mathbf{\hat{Y}}_\tau^s\right|.
\end{equation}
The training procedure of DualSTN can be summarized in Algorithm~\ref{algo1}. For each iteration, we randomly sample a subgraph and its corresponding sensor readings to train the model. Note that we adopt the same subgraph and node division for each batch to simplify the implementation.  

\begin{algorithm}[t]
\caption{Training Procedure of DualSTN}
\begin{algorithmic}[1]
\REQUIRE graph $\mathcal{G}=(\mathcal{V}, \mathcal{E})$, sensor readings of training set $\mathcal{X}$, time window $T$; initialized model $\operatorname{DualSTN}()$.
\ENSURE optimized learnable weights.
\FOR{$i=1 \to Num\_Iteration$}
\STATE Initialize batch list $\mathbf{X_b}=[ ]$, $\mathbf{Y_b}=[ ]$.
\\\COMMENT{Sampling Graph}
\STATE $\mathcal{G}_s$, $\mathbf{X_s}$, $\mathbf{Y_s}$ = Graph\_Sampling($\mathcal{G}$, $\mathcal{X}$).
\FOR{$j=1 \to Batch\_Size$}
\STATE Randomly choose a start time $t$.
\STATE Append $\{\mathbf{X_s}\}_{t:t+T}$ to $\mathbf{X_b}$.
\STATE Append $\{\mathbf{Y_s}\}_{t+T}$ to $\mathbf{Y_b}$.
\ENDFOR
\ENDFOR
\STATE $\mathbf{\hat{Y}_b}^{s}$, $\mathbf{\hat{Y}_b}^{l} = \operatorname{DualSTN}(\mathcal{G}_s, \mathbf{X_b})$.
\STATE Compute $\operatorname{MAE}(\mathbf{\hat{Y}_b}^{s}, \mathbf{Y_b})+\operatorname{MAE}(\mathbf{\hat{Y}_b}^{l}, \mathbf{Y_b})$ and derive the gradients.
\STATE Update learnable weights using the optimizer.
\end{algorithmic}
\label{algo1}
\end{algorithm}

\section{Experiments}
\subsection{Experimental Settings}
\subsubsection{Datasets}
We evaluate the performances of DualSTN on four real-world spatiotemporal datasets in diverse application scenarios: 1) \textbf{METR-LA}\footnote{\url{https://github.com/liyaguang/DCRNN}}~\cite{li2017diffusion}: trafﬁc speed dataset collected from 207 sensors in the highway of Los Angeles from 01/05/2012 to 30/06/2012. 2) \textbf{PeMS-Bay}\footnote{\url{https://github.com/liyaguang/DCRNN}}~\cite{li2017diffusion}: a trafﬁc speed dataset collected by California Transportation Agencies containing 325 sensors in the Bay Area from 01/01/2017 to
13/05/2017. 3) \textbf{NREL}\footnote{\url{https://www.nrel.gov/grid/solar-power-data.html}}~\cite{sengupta2018national}: energy datasets provided by the National Renewable Energy Laboratory and we choose a subset of Alabama Solar Power Data. The dataset contains 137 photovoltaic power plant readings collected in 2006. 4) \textbf{BJ-Air}\footnote{\url{https://www.biendata.xyz/competition/kdd_2018/}}: air quality index dataset from 35 air quality stations in Beijing and we consider the PM2.5 observations. 

We construct the pre-defined adjacency matrix $\mathbf{{A}}$ based on either road network distance or geospatial distance $dist(v_i, v_j)$. The road network distance is available in the dataset and we compute geospatial distance using Haversine
formula, given the longitude and latitude:
\begin{equation}
\begin{split}
    dist(v_i, v_j) =2 r \arcsin \left(\sin ^{2}\left(\frac{\varphi_{j}-\varphi_{i}}{2}\right)+ \right.\\
    \left.\cos \left(\varphi_{i}\right) \cos \left(\varphi_{j}\right) \sin ^{2}\left(\frac{\lambda_{j}-\lambda_{i}}{2}\right)\right)^{\frac{1}{2}},
\end{split}
\end{equation}
where $r=6371$ is the radius of the earth, $(\varphi_{i}, \lambda_{i})$ means the longitude and latitude of the sensor $v_i$.
Then the Gaussian kernel method~\cite{shuman2013emerging} is applied:
\begin{equation}
    \mathbf{A}_{i,j}=\exp(-\frac{1}{2} \frac{dist(v_i, v_j)^2}{\sigma^2}),
\end{equation}
where $\sigma$ is the standard deviation. In the case of the directed graph that contains bi-directional adjacency matrices $\mathbf{A}_f$ and $\mathbf{A}_b$, we use the same network to model them which is equivalent to $\mathbf{A}=(\mathbf{A}_f+\mathbf{A}_b)/2$. We summarize the statistics of datasets in Table~\ref{table_dataset}.

\begin{table}[t]
\caption{Dataset statistics. \#: the number. Traffic Spd: traffic speed. Rn-dis/Geo-dis: road-network/geospatial distance.}
\label{table_dataset}
  \scalebox{0.9}{
  \begin{threeparttable}
  \begin{tabular}[width=0.78\linewidth]{ccccc}
    \toprule
      Dataset & METR-LA & PeMS-Bay & NREL & BJ-Air\\
      \midrule
      Category & Traffic Spd & Traffic Spd & Solar Energy & Air Quality\\
      Adjacency Matrix & Rn-Dis & Rn-Dis & Geo-Dis & Geo-Dis\\
      \# Sensors & 207 & 325 & 137 & 35\\
      \# Time points & 34,272 & 52,116 & 105,120 & 10,228\\
      Frequency & 5-min & 5-min & 5-min & 1-hour\\
      Mean & 58.45 & 62.62 & 15.96 & 60.99 \\
      Standard Deviation & 13.08 & 9.58 & 9.86 & 65.31\\
  \bottomrule
\end{tabular}
\end{threeparttable}
}
\end{table}

\subsubsection{Baseline Methods}
We compare the performances of our model with 7 baselines as follows:
\begin{itemize}
    \item \textbf{KNN}: K-nearest neighbors interpolate readings of unknown locations by averaging the $k$-nearest sensors in the spatial dimension.
    \item \textbf{IDW}: Inverse distance weighting utilizes distances between nodes to calculate a weighted average of available nodes for each unknown location~\cite{lu2008adaptive}. 
    \item \textbf{OKriging}: Ordinary kriging is a classic statistical interpolation method based on the geospatial locations of the sensors and Gaussian processes~\cite{cressie2015statistics}. We evaluate the performance of OKriging using the package PyKrige\footnote{\url{https://geostat-framework.readthedocs.io/projects/pykrige/en/stable}}. Note that OKriging is not applicable for road network distance, so we just report the performances on NREL and BJ-Air datasets.
    \item \textbf{GLTL:}\footnote{\url{https://roseyu.com/code.html}} Greedy Low-rank Tensor Learning is a low-rank tensor learning framework for spatiotemporal data co-kriging and forecasting which handles various properties in the data. Moreover, a fast greedy algorithm is proposed to learn the tensor efﬁciently. 
    \item \textbf{KCN}, \textbf{KCN-SAGE}\footnote{\url{https://github.com/tufts-ml/KCN}}: Kriging Convolutional Network first searches $k$-nearest neighbors for a target location. Then, it constructs a graph structure for the K + 1 nodes as inputs of GNNs to interpolate signals~\cite{appleby2020kriging}. KCN-SAGE is a variant based on graph sampling and aggregating~\cite{hamilton2017inductive}. 
    \item \textbf{IGNNK}\footnote{\url{https://github.com/Kaimaoge/IGNNK}}: Inductive Graph Neural Network Kriging is a state-of-the-art model trained in an inductive approach~\cite{wu2020inductive}. It regards sequential reading as the feature of a GNN to learn spatial and temporal relations.
    \item \textbf{SATCN}\footnote{\url{https://github.com/Kaimaoge/SATCN}}: Spatial Aggregation and Temporal Convolution Network contains a spat of spatial aggregators using signals' statistic features for spatial modeling. Meanwhile, TCNs are leveraged to capture temporal dependencies~\cite{wu2021spatial}. 
\end{itemize}

\subsubsection{Evaluation Metrics}
We utilize three criteria to evaluate models: the root mean square error (RMSE), the mean absolute error (MAE), and the mean absolute percentage error (MAPE). All of them are frequently used in regression problems:
\begin{equation}
    \operatorname{RMSE}=\sqrt{\frac{1}{N} \sum_{i \in N}\left(y^{i}-\hat{y}^i\right)^{2}},
\end{equation}
\begin{equation}
    \operatorname{MAE}=\frac{1}{N} \sum_{i \in N}\left|y^{i}-\hat{y}^i\right|,
\end{equation}
\begin{equation}
    \operatorname{MAPE}=\frac{1}{N} \sum_{i \in N} \frac{\left|{y}^{i}-{\hat{y}}^i\right|}{{y}^i},
\end{equation}
where ${y}^i$ is the ground truth signal of a sensor and $\hat{y}^i$ denotes inferred signals.
\subsubsection{Implementation Details}

\begin{table*}
\center
\caption{Performance comparison and the number of learnable parameters for different methods on four datasets. K: thousand.}
\label{table1}
  \begin{threeparttable}
  \begin{tabular}[width=1.\linewidth]{lccccccc}
    \toprule
      \multirow{2}*{Model} & \multicolumn{3}{c}{METR-LA} & \multicolumn{3}{c}{PeMS-Bay} & \multirow{2}*{\#Params}\\
      \cmidrule(r){2-4}  \cmidrule(r){5-7}
      & MAE & RMSE & MAPE & MAE & RMSE & MAPE & \\
      \midrule
      KNN & 7.65\tpm{0.00} & 11.20\tpm{0.00} & 0.183 & 6.45\tpm{0.00} & 12.16\tpm{0.00} &  0.112 & --\\
      IDW & 7.78\tpm{0.00} & 11.53\tpm{0.00} & 0.188 & 6.38\tpm{0.00} & 11.78\tpm{0.00} & 0.103  & -- \\
      OKriging & -- & -- & -- & -- & -- & --  & -- \\
      GLTL & 7.71\tpm{0.00} & 11.03\tpm{0.00}  & 0.186 & 5.20\tpm{0.00} & 8.90\tpm{0.00}  & 0.0983 & -- \\
      KCN & 7.19\tpm{0.04} & 10.55\tpm{0.07} & 0.183 & 4.70\tpm{0.07} & 8.12\tpm{0.06} &  0.095  & 18K\\
      KCN-SAGE & 7.06\tpm{0.04} & 10.05\tpm{0.07} & 0.179 & 4.38\tpm{0.06} & 7.50\tpm{0.07} &  0.092  & 15K\\
      IGNNK & 6.93\tpm{0.09} & 10.59\tpm{0.05} & 0.175 & 4.06\tpm{0.06} & 7.12\tpm{0.08} & 0.090  & 45K\\
      SATCN & 7.03\tpm{0.05} & 10.51\tpm{0.07} & 0.176 & 4.04\tpm{0.03} & 6.96\tpm{0.02} &  0.092  & 66K\\
      DualSTN (ours) & \textbf{6.73\tpm{0.03}} & \textbf{9.93\tpm{0.02}} & \textbf{0.171} & \textbf{3.95\tpm{0.06}} &\textbf{6.68\tpm{0.04}} & \textbf{0.088} & \textbf{12K}\\
 \midrule
       \multirow{2}*{Model} & \multicolumn{3}{c}{NREL} & \multicolumn{3}{c}{BJ-Air} & \multirow{2}*{\#Params}\\
      \cmidrule(r){2-4}  \cmidrule(r){5-7}
      & MAE & RMSE & MAPE & MAE & RMSE & MAPE & \\
      \midrule
      KNN & 3.28\tpm{0.00} & 4.62\tpm{0.00}  & 1.031 & 18.35\tpm{0.00} & 29.41\tpm{0.00}  & 1.044 & --\\
      IDW & 3.08\tpm{0.00} & 4.46\tpm{0.00}  & 0.932 & 17.80\tpm{0.00} & 28.21\tpm{0.00} & 0.921 & -- \\
      OKriging & 2.81\tpm{0.00} & 4.23\tpm{0.00}  & 0.855 & 17.98\tpm{0.00} & 28.62\tpm{0.00}  & 0.954 & -- \\
      GLTL & 3.20\tpm{0.00} & 4.49\tpm{0.00}  & 0.937 & 16.33\tpm{0.00} & 26.94\tpm{0.00}  & 0.933 & -- \\
      KCN & 1.71\tpm{0.08} & 2.90\tpm{0.08} & 0.739 & 14.63\tpm{0.20} & 25.39\tpm{0.32} & 0.743   & 18K\\
      KCN-SAGE & 1.65\tpm{0.05} & 2.84\tpm{0.07} & 0.701 & 14.40\tpm{0.26} & 24.98\tpm{0.27} &  0.725  & 15K\\
      IGNNK & 1.53\tpm{0.06} & 2.75\tpm{0.08} & 0.682 & 13.79\tpm{0.22} & 24.95\tpm{0.29}  & 0.651  & 45K\\
      SATCN & 1.69\tpm{0.05} & 2.90\tpm{0.02} & 0.735 & \textbf{12.93\tpm{0.11}} & 23.39\tpm{0.22} &  0.560  & 66K\\
      DualSTN (ours) & \textbf{1.49\tpm{0.03}} & \textbf{2.69\tpm{0.02}}  & \textbf{0.643} & {13.01\tpm{0.20}} & \textbf{23.30\tpm{0.28}}  & \textbf{0.554} & \textbf{12K}\\
  \bottomrule
\end{tabular}
\end{threeparttable}

\end{table*}

Our DualSTN and other deep-learning baselines are implemented with PyTorch 1.7 and trained on a Quadro RTX 6000 GPU. 
We use a historical time window $T=25$ to infer the real-time readings in which the time window for short-term learning is $t_s=3$. The skip step of GRU $t_k$ equals 4, i.e., we feed readings of sensors into the network every 4 steps.
For the hyperparameters, we set decay rates $\rho$ and $\lambda$ to 1, saturation rate $\alpha$ to 2, and impact factors $\gamma$, $\mu$ to 0.1 and 0.9. The activation function $\phi$ is ReLU. The number of layers for JST-GAT is 3 and the size of hidden states for graph GRU is 16. 
All learnable parameters are initialized with the Xavier~\cite{glorot2010understanding}. The model is trained by the Adam~\cite{kingma2014adam} optimizer and the learning rate of $10^{-3}$. Note that we keep the same settings for all datasets, verifying the generalization ability of our model. For the dataset division, we use the first 70\% of the time frames to train models and validate or test models using the following 20\% and 10\%, respectively. For the sensor division, we manually leave 50\% of the sensors out for testing, dubbed testing sensors, and the remaining 50\% as training sensors. At each epoch, we randomly mask 50\% of the training sensors as unknown locations. When evaluating models, we use all training sensors to interpolate all testing sensors. To ensure a fair comparison, we use the same division of training and testing sensors and train each deep learning model five times independently to report the average results and the standard deviations.

\subsection{Model Comparison}
In this section, we compare the performances of our DualSTN with all baselines, and the results are summarized in Table~\ref{table1}. We observe that statistical methods have worse results than data-driven approaches. This is chiefly because they are designed to capture linear or Gaussian relations, which is suboptimal to measure complex dependencies. 
For deep learning methods, as they could learn complicated non-linear spatiotemporal dependencies from data, we find that even the spatial method KCN outperforms GLTL which considers both spatial and temporal dependencies. 
Meanwhile, IGNNK outperforms KCN because IGNNK adopts n-hop GNNs and learns from temporal information. 
For our model, we observe that it outperforms all baselines on LETR-LA, PeMS-Bay, and NREL and competitive results on the BJ-Air dataset. 
It results from that DualSTN could learn spatiotemporal relationships simultaneously and the decoupled design makes the model easier to distinguish long- and short-term patterns. In addition, DualSTN also has fewer parameters  than other models, which also demonstrates its learning ability. The reason might be that our model uses a single module to learn spatial and temporal relations. Thus, we do not need to stack many layers to enlarge the receptive field, reaching nodes with far distances. 
For the BJ-Air dataset, both our DualSTN and SATCN achieved commensurate performance and surpass other baselines. SATCN designs special aggregators like the standard deviation aggregator, which might be beneficial for this dataset with a large deviation. However, DualSTN has five times fewer parameters than SATCN. 
Furthermore, we notice that on the BJ-Air dataset, the results of the three deep learning models have larger variations, which means that the training process is not stable. We conjecture the limited number of sensors and the large standard deviation of readings cause difficulties in learning general spatiotemporal dependencies.

\begin{figure}[!b]
\centering
\subfloat{
\begin{minipage}[t]{0.5\linewidth}
\centering
\includegraphics[width=0.98\linewidth]{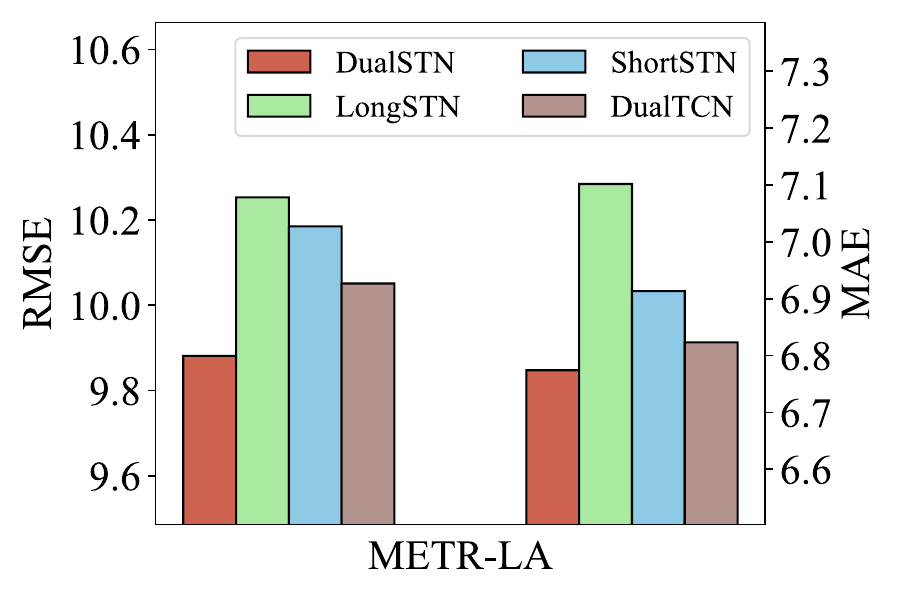}
\end{minipage}%
}%
\subfloat{
\begin{minipage}[t]{0.5\linewidth}
\centering
\includegraphics[width=0.98\linewidth]{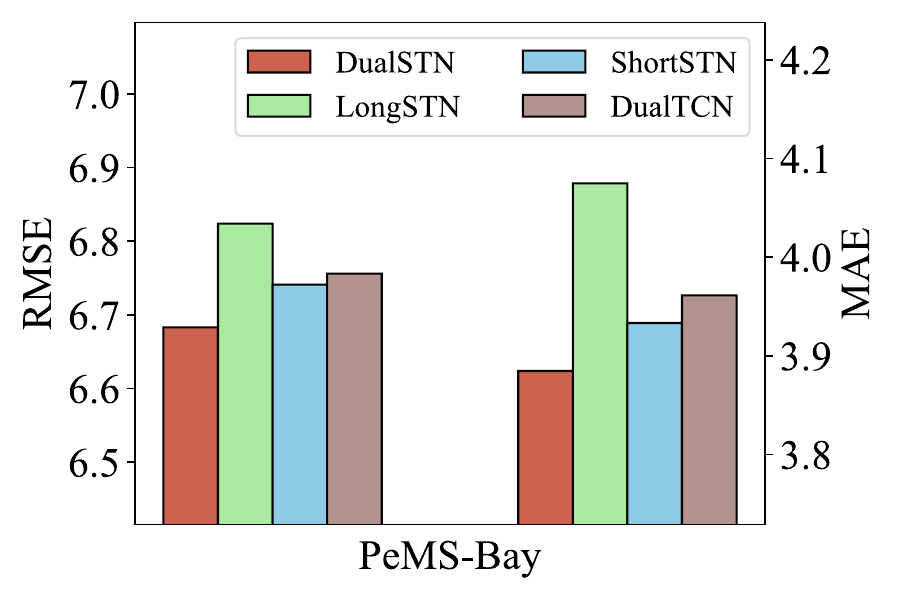}
\end{minipage}
}

\subfloat{
\begin{minipage}[t]{0.5\linewidth}
\centering
\includegraphics[width=0.98\linewidth]{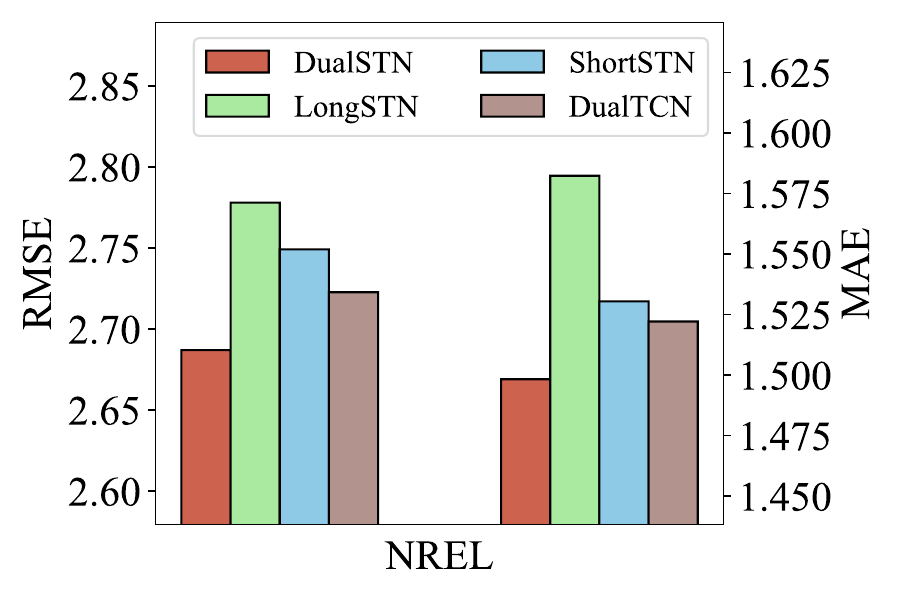}
\end{minipage}
}%
\subfloat{
\begin{minipage}[t]{0.5\linewidth}
\centering
\includegraphics[width=0.98\linewidth]{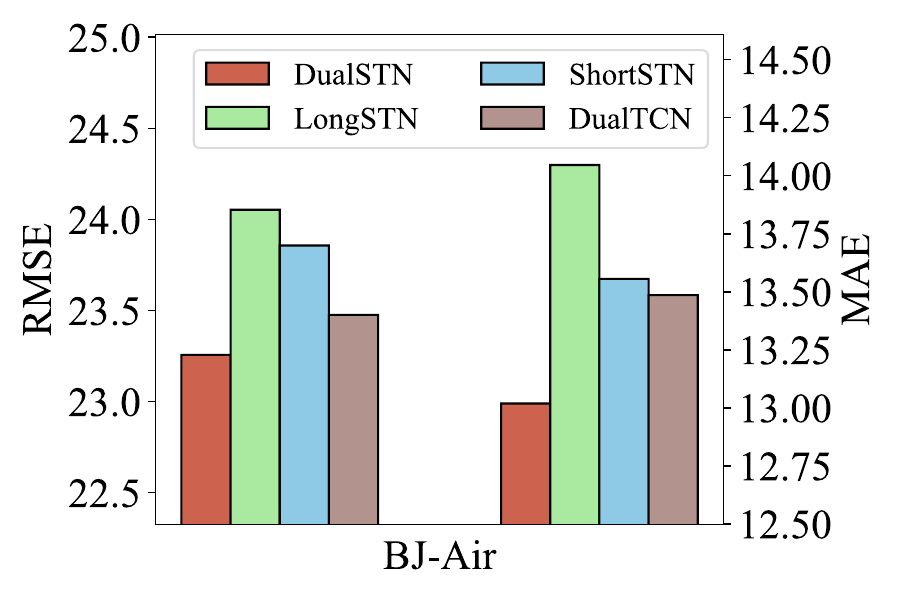}
\end{minipage}
}%
\centering
\caption{Ablation studies. The DualSTN consistently achieves the best RMSE and MAE results against other variants.}
\label{figure_ablation1}
\end{figure}

\subsection{Ablation Study}
Our model is largely built upon two motivations (i.e., decoupling long- short-term patterns and joint spatiotemporal learning). A natural question is whether they are effective. In the section, we implement three variants of DualSTN to verify the effectiveness of components described as follows:

\begin{itemize}
    \item \textbf{LongSTN}: This variant removes the short-term learning module and only contains the skip graph gated recurrent unit. The inference results are outputs of the GRU's last recurrent step.
    \item \textbf{ShortSTN}: This variant removes the skip GRU and remains the joint spatiotemporal attention graph network which only takes short-term frames as the inputs to infer unknown locations.
    \item \textbf{DualTCN}: Our model uses attention blocks to learn temporal dependencies in the short-term module and another intuitive opinion is leveraging TCNs. Thus, this variant replaces the attention module with the temporal convolution network (TCN) that follows the same structure as~\cite{wu2019graph} and the long-term module remains the same.
\end{itemize}

We evaluate the performance of three variants on four datasets and illustrate the results in Figure~\ref{figure_ablation1}. We find that ShortSTN performs better than LongSTN while both are much worse compared to DualSTN. These observations demonstrate the following conclusions. 1) Sensor readings are more relevant to short-term patterns but still follow the trend of long-term ones. 2) Decoupling long- and short-term learning improves performance as they provide information from different perspectives. For the TCN variant, we observe that DualSTN has a better performance compared to DualTCN and argue that this is because the joint attention module explicitly takes joint spatiotemporal relations into consideration, which reduces the challenge of modeling complex relations.

\begin{figure*}[t]
  \centering
  \includegraphics[width=0.9\linewidth]{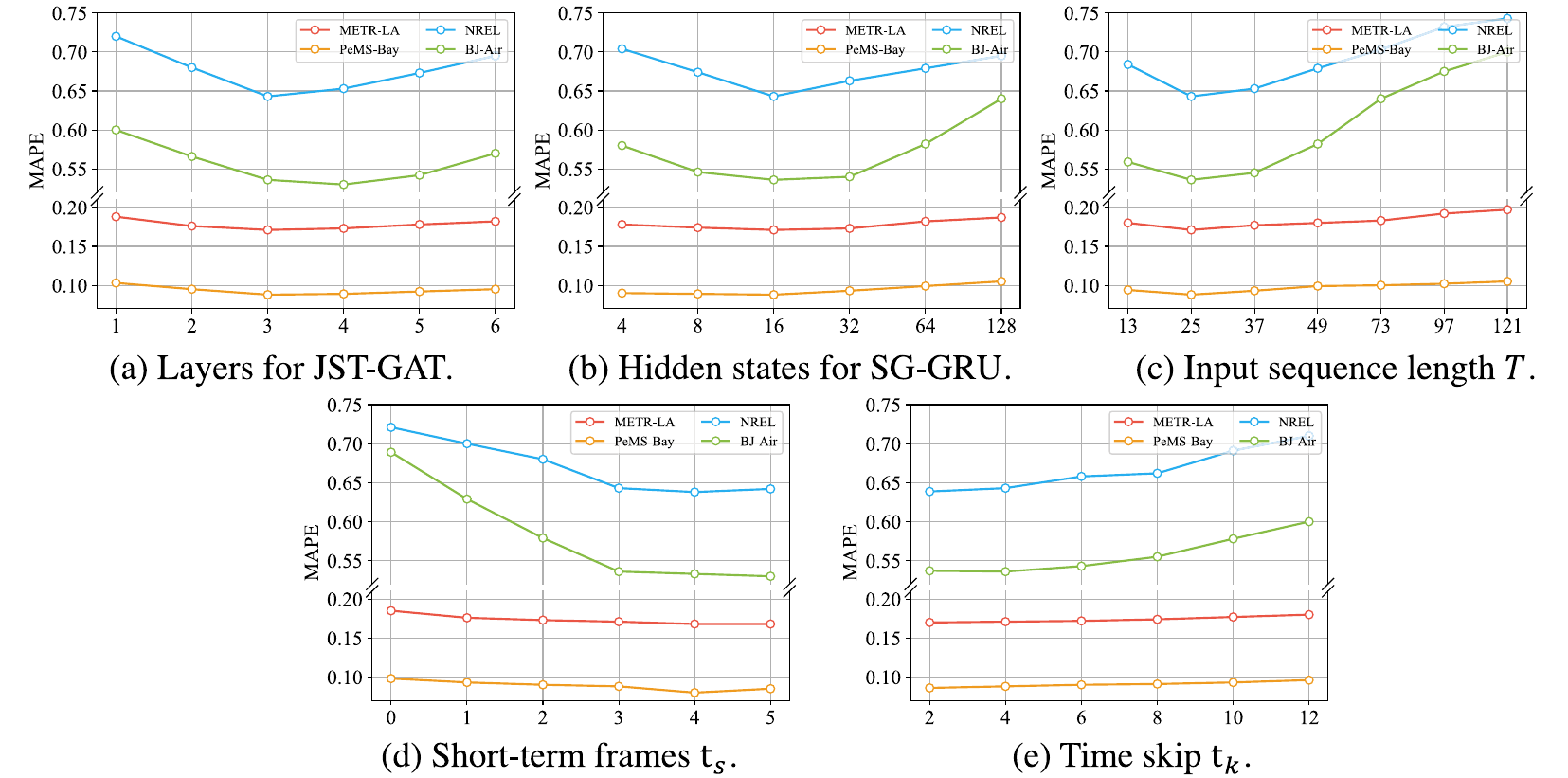}
  \caption{Parameters study among different variants on METR-LA, PeMS-Bay, NREL, and BJ-Air datasets.}
    \label{figure_ablation2}
\end{figure*}

\subsection{Hyperparameter Study}
In this section, we study the performance of our DualSTN under different hyperparameter settings and report MAPE results. In each study, we modify the setting of corresponding hyperparameters and keep others unchanged. All the experiments are conducted on four datasets. 

\subsubsection{Effects of number of layers of JST-GAT}
We adjust the number of layers $L$ in the joint spatiotemporal attention module and report the results in Figure~\ref{figure_ablation2}(a). The model performances first become better and achieve best at 4 layers for the BJ-Air dataset and 3 layers for the rest. Then, MAPE results remain stable or start to increase slightly. According to these observations, we uniformly keep the number of layers as 3 to reduce the computational cost. 

\subsubsection{Effects of size of hidden states of SG-GRU}
We change the size of the graph GRU's hidden states from 4 to 128. As shown in Figure~\ref{figure_ablation2}(b), we discover that the performances increase until reaching the size of 16 or 32 for four datasets. Then they start to decrease, indicating that the model tends to overfit. Accordingly, we set the size to 16.

\subsubsection{Effects of time window $T$}
We keep the skip step in the graph GRU $t_k=4$ and adjust the input time window $T$ to evaluate GRU's capability of learning long-term patterns. As expected in Figure~\ref{figure_ablation2}(c), a longer input sequence cannot guarantee better results. Instead, the model crashes over a long time window because of its gradient vanishing problem of GRU on long-term modeling. In this scenario, the encoded long historical features become noises of hidden states, impeding the model performance.

\subsubsection{Effects of number of short-term frames $t_s$}
Next, we modify the number of short-term frames $t_s$. As illustrated in Figure~\ref{figure_ablation2}(d), as $t_s$ increases, the performance ﬁrst increases fast and then levels off. As a larger $t_s$ uses frames modeled by SG-GRU, the delicate learning module JST-GAT is redundant to capture them again. To this end, we set $t_s$ to 3 and leverage the graph GRU for learning these patterns.

\subsubsection{Effects of time skip $t_k$}
Finally, we keep $t_s=3$, $T=25$ and adjust time skip $t_k$ to evaluate its influence. As shown in Figure~\ref{figure_ablation2}(e), the results consistently decrease and especially, this decrease speeds up as $t_k$ increases. The reason is that as the time span becomes too large, the temporal dependencies between input frames become sparse that the model cannot capture. Thus, we choose $t_k$ to 4 to ensure no frame overlap between the short-term and long-term modules which also saves running time.

From the study, we observe that DualSTN is able to achieve satisfying performances over all datasets using the same hyperparameter setting. This means that our model is insensitive to different application domains, which relieves the demand for hyperparameter searching and is beneficial to real-world deployment. It could be caused by the fewer learnable parameters and the learning effectiveness of the model.

\subsection{Case Study}
\subsubsection{Long- and short-term patterns inconsistency}
\begin{figure}[!h]
\centering
\subfloat{
\begin{minipage}[t]{1\linewidth}
\centering
\includegraphics[width=1.\linewidth]{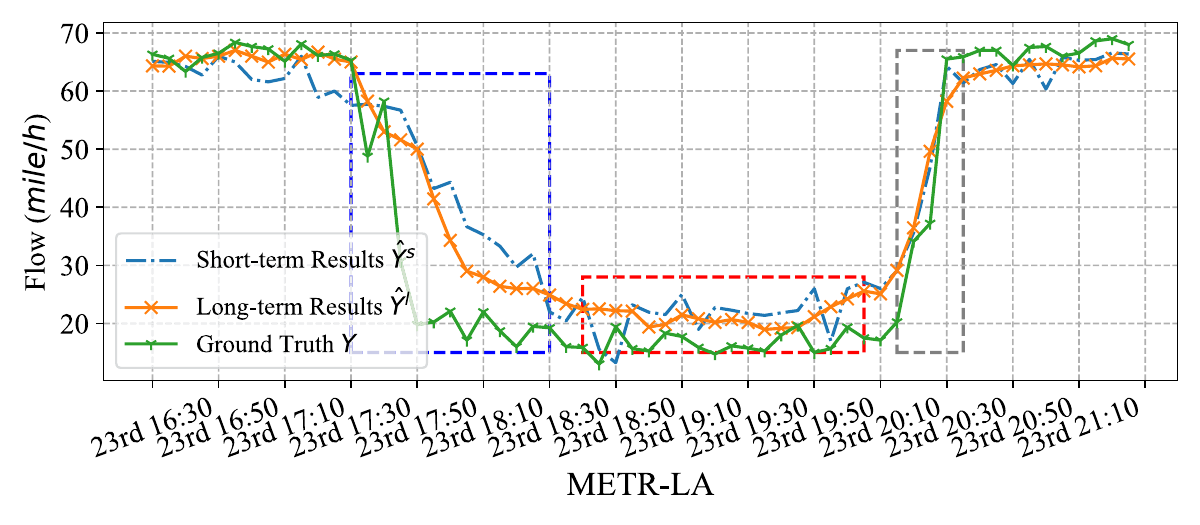}
\end{minipage}%
}%

\subfloat{
\begin{minipage}[t]{1\linewidth}
\centering
\includegraphics[width=1.\linewidth]{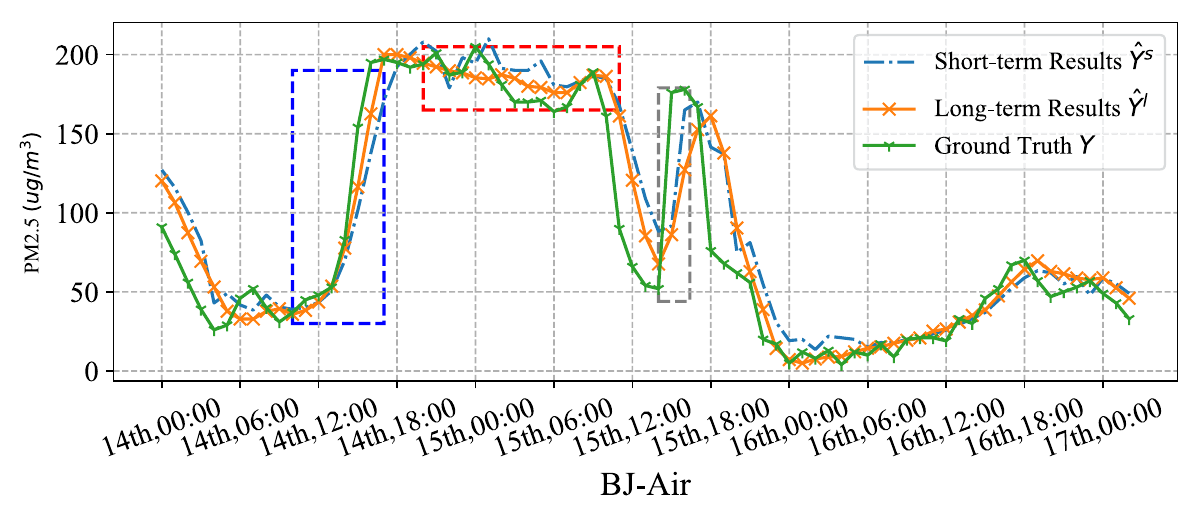}
\end{minipage}
}
\caption{Visualizations of short-term and final long-term inference results compared to the ground truth on METR-LA and BJ-Air datasets. The time starts from 16:30, 23/06/2023, and 0:00, 14/01/2018, respectively.}
\label{figure_case2}
\end{figure}

To study how our DualSTN captures the long- and short-term patterns, we visualize the short-term results $\mathbf{\hat{Y}}^s$ and the final results $\mathbf{\hat{Y}}^l$ that integrates information of both terms. Figure~\ref{figure_case2} shows results and ground truth of META-LA from 16:30, 23/06/2023, and the BJ-Air dataset from 14/01/2018, in which we have three observations. 
1) In the red boxes, the truth signals fluctuate while having a flat trend. The short-term inferred signals, aligning with the ground truth, also oscillate as the JST-GAT only focuses on short-term patterns. Then by involving trends from long-term patterns, the final outputs become stable. 2) In the blue boxes, signals follow an upward or downward trend. The long-term outputs are accurate compared to the short-term results. This suggests the tendency is important in this scenario and SG-GRU could capture it precisely. 3) In the gray boxes where a sudden change in readings happens, the short-term outputs outperform the long-term results. This is because the historical tendency in SG-GRU does not tally with this sudden change. These discoveries mean that our model handles short- and long-term patterns, and JST-GAT and SG-GRU can contribute to the final results in different aspects.  

\subsubsection{Joint spatiotemporal Dependencies}
\begin{figure}[h]
  \centering
  \includegraphics[width=1.\linewidth]{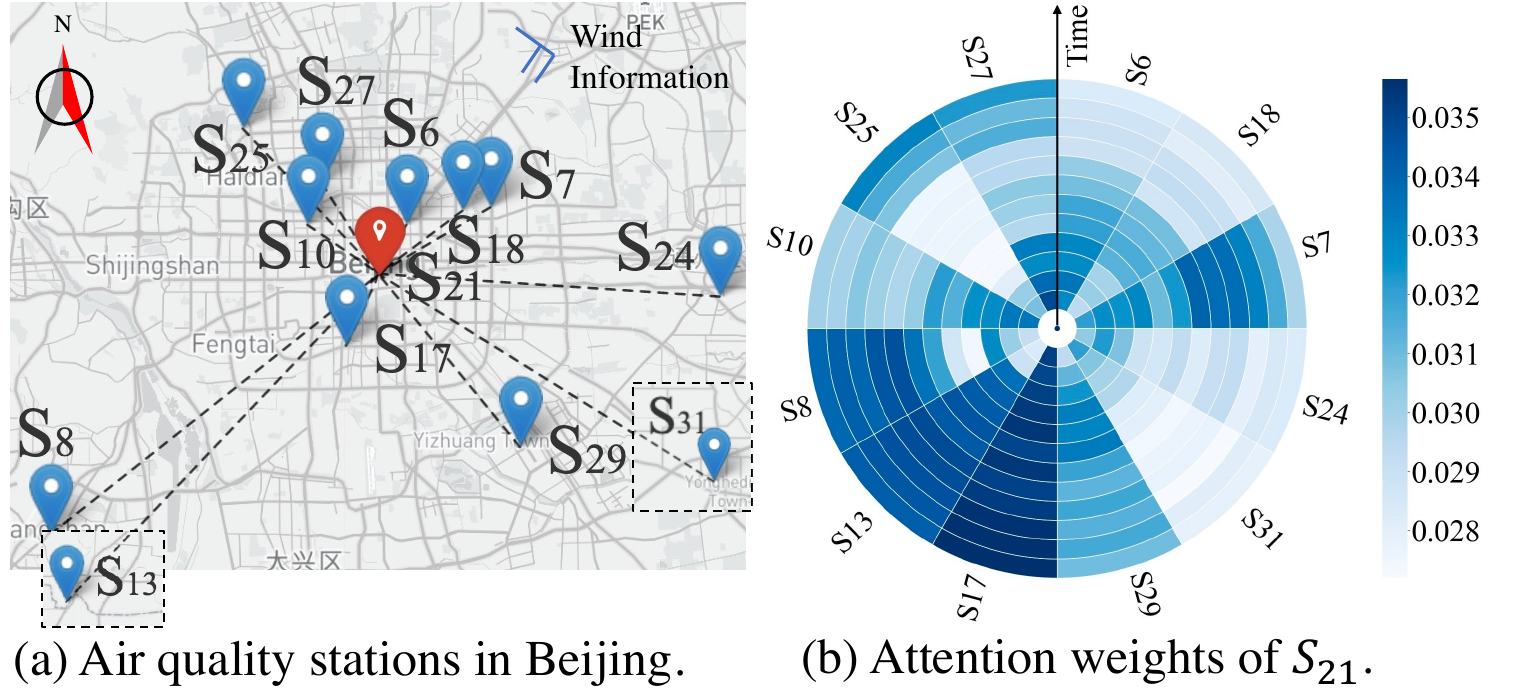}
  \caption{(a) Locations of stations, where only the discussed sensors are shown. $S_{21}$ is the target sensor to compute attention weights. Note that stations $S_{13}$, $S_{31}$ are extremely far from $S_{21}$. (b) Attention scores of station $S_{21}$.}
    \label{figure_case}
\end{figure}

The attention block in our model provides interpretability by indicating the dependencies between two nodes. As the motivation here is to capture joint spatiotemporal relations, we conduct a case study using the BJ-Air dataset from 0:00 to 12:00 on 03/09/2017 and visualize the attention weights of an inferred location to investigate this ability. 
For succinctness, we use a center station $S_{21}$ as the target location and compute attention weights between its signals and those of equably sampled 13 sensors at different times. 
Figure~\ref{figure_case}(a) illustrates their locations and 
Figure~\ref{figure_case}(b) shows the attention scores of 12 consecutive time steps, where stations are drawn according to their relative orientations to $S_{21}$. Given the southeastern condition, we have the following conclusions.
1) Stations $S_{8}$, $S_{13}$, and $S_{17}$ have larger weights, meaning that these stations contribute more to the inference of $S_{21}$.
2) The pre-defined adjacency matrix could reflect the spatial relations regarding geographical information, as near stations have a large influence (e.g., $S_{29}$ and $S_{31}$ to $S_{21}$).
3) However, it cannot guarantee genuine spatiotemporal dependencies. For instance, $S_{13}$ is excessively far away from $S_{21}$ compared to $S_{18}$ but has an even larger impact. In this situation, the dynamic dependencies become dominant. 
4) The weights of $S_8$ increase over time, mightily due to the change of wind speed. This means that joint spatiotemporal attention is capable of capturing relations across time and space. 
These observations verify that our model captures both static spatial relations, dynamic implicit dependencies, and joint spatiotemporal relations simultaneously even without knowing possible external factors. This compelling learning ability also provides the possibility of applying the module to other tasks that require delicate spatial and temporal modeling, such as video object segmentation~\cite{huang2021scribble,zhang2018spftn}.

\section{Conclusions and Future Works}
In this paper, we propose a novel DualSTN model for spatiotemporal inference.
To better learn the short- and long-term patterns, we decouple the model into two components: JST-GAT and SG-GRU. 
The first aims to capture delicate short-term joint spatiotemporal correlations while the second network focuses on long-term patterns by a time skip strategy.
Extensive experiments on four real-world applications suggest that our DualSTN offers state-of-the-art performances against previous baselines. Further evaluations also justify the effectiveness of the modules as well as the interpretation ability brought by attention mechanisms.

In the future, we can explore ways to speed up the inference without losing performance, as we find that SG-GRU consumes a long inference time due to its recurrent structure.
Then, the idea of joint attention can be transferred to the forecasting task for better capturing spatiotemporal relations. 
In addition, more complex models can be proposed to integrate forecasting and inference as a unified task.  

\section*{Acknowledgments}
This research is supported by Singapore Ministry of Education Academic Research Fund Tier 2 under MOE's official grant number T2EP20221-0023. It is also supported by Guangzhou Municipal Science and Technology Project 2023A03J0011.

%

\bibliographystyle{IEEEtran}
\bibliography{ref}


\newpage
\begin{IEEEbiography}
[{\includegraphics[width=1in,height=1.25in,clip,keepaspectratio]{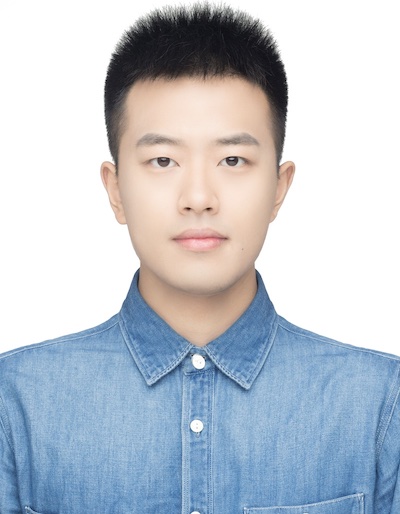}}]{Junfeng Hu} received the B.E. degree from the School of Big Data \& Software Engineering, Chongqing, China, in 2020 and received the Master degree from the National University of Singapore, in 2022. He is currently a Ph.D. student in School of Computing at National University of Singapore. His research interests include spatio-temporal data mining and graph neural networks. He has published papers at conferences such as KDD, ECAI, ECML, IJCNN, and ICME.
\end{IEEEbiography}

\begin{IEEEbiography}
[{\includegraphics[width=1in,height=1.25in,clip,keepaspectratio]{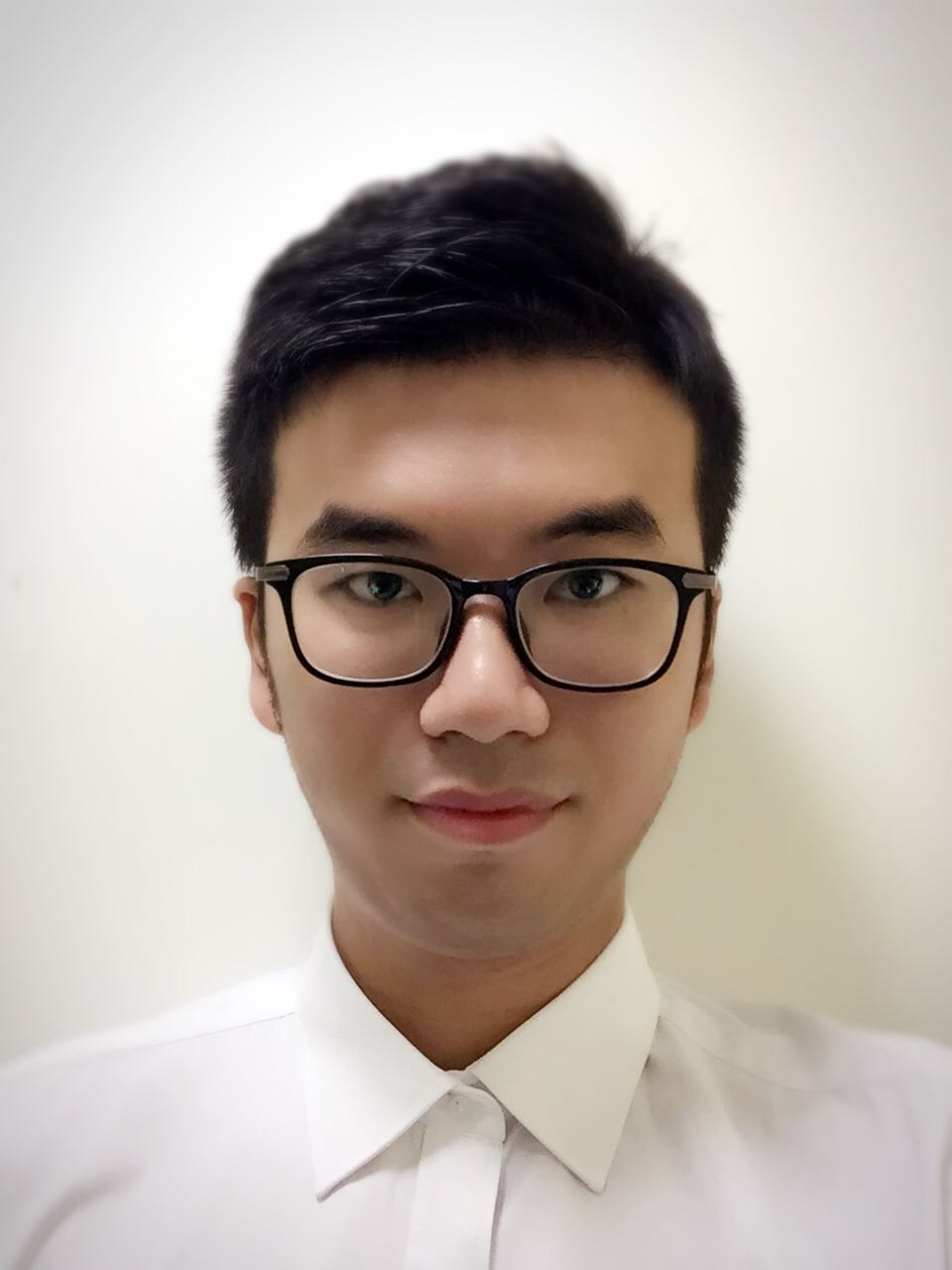}}]{Yuxuan Liang} is an Assistant Professor at Intelligent Transportation Thrust, Hong Kong University of Science and Technology (Guangzhou). He is currently working on the research, development, and innovation of spatio-temporal data mining and AI, with a broad range of applications in smart cities. Prior to that, he obtained his PhD degree at NUS. He published over 40 peer-reviewed papers in refereed journals and conferences, such as KDD, WWW, NeurIPS, ICLR, ECCV, AAAI, IJCAI, Ubicomp, and TKDE. Those papers have been cited over 2,000 times (Google Scholar H-Index: 21). He was recognized as 1 out of 10 most innovative and impactful PhD students focusing on data science in Singapore by Singapore Data Science Consortium (SDSC).

\end{IEEEbiography}

\begin{IEEEbiography}
[{\includegraphics[width=1in,height=1.25in,clip,keepaspectratio]{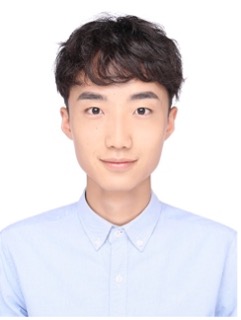}}]{Zhencheng Fan} is a Ph.D. candidate at the University of Technology Sydney, Australia, specializing in memristor-based neural networks, data mining, and computer vision. He received his Bachelor's degree in Software Engineering from Chongqing University, China in 2020. His work has been published in conferences including KDD, ECAI, and KSEM.
\end{IEEEbiography}

\begin{IEEEbiography}
[{\includegraphics[width=1in,height=1.25in,clip,keepaspectratio]{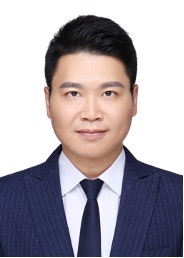}}]{Li Liu} is a professor at Chongqing University. He is also serving as a Senior Research Fellow of School of Computing at the National University of Singapore. Li received his Ph.D. in Computer Science from the Université Paris-sud XI in 2008. His research interests are in pattern recognition, data analysis, and their applications on human behaviors. He aims to contribute in interdisciplinary research of computer science and human related disciplines. Li has published widely in conferences and journals with more than 100 peer-reviewed publications. Li has been the Principal Investigator of several funded projects from government and industry.
\end{IEEEbiography}

\begin{IEEEbiography}
[{\includegraphics[width=1in,height=1.25in,clip,keepaspectratio]{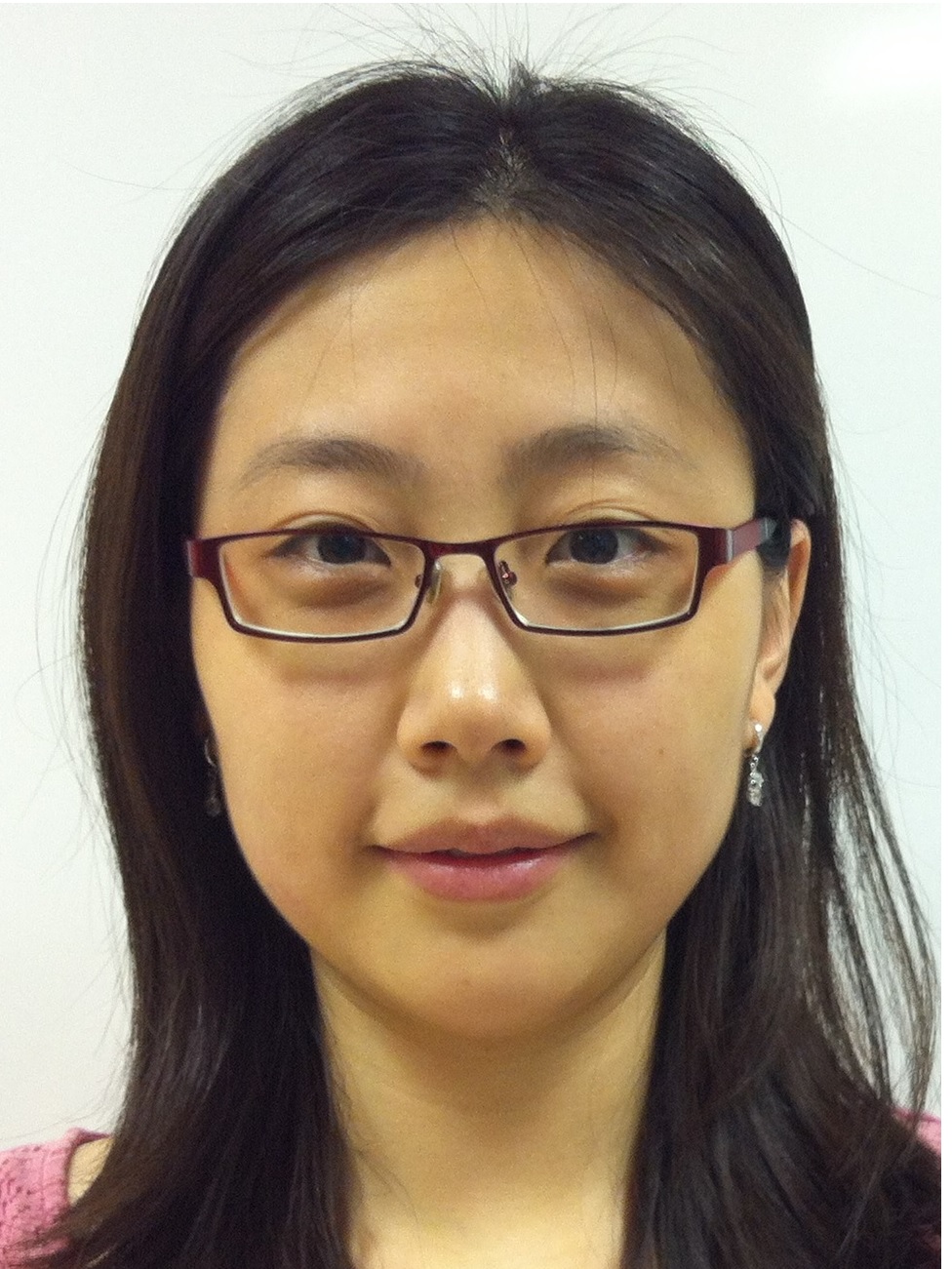}}]{Yifang Yin} received the B.E. degree from the Department of Computer Science and Technology, Northeastern University, Shenyang, China, in 2011, and received the Ph.D. degree from the National University of Singapore, Singapore, in 2016. She is currently a scientist at Institute for Infocomm Research, A*STAR, Singapore. She also holds an adjunct faculty position at IIIT-Delhi. Before joining A*STAR, she worked as a senior research fellow with the Grab-NUS AI Lab at the National University of Singapore. She also worked as a Research Intern at the Incubation Center, Research and Technology Group, Fuji Xerox Co., Ltd., Japan, from October, 2014 to March, 2015. Her research interests include machine learning, spatiotemporal data mining, and multimodal analysis in multimedia.
\end{IEEEbiography}

\begin{IEEEbiography}
[{\includegraphics[width=1in,height=1.25in,clip,keepaspectratio]{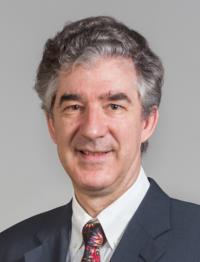}}]{Roger Zimmermann} received his M.S. and Ph.D. degrees from the University of Southern California (USC) in 1994 and 1998, respectively. He is currently a Professor with the Department of Computer Science at the National University of Singapore (NUS) and a key investigator with the Grab-NUS AI Lab.  He previously was a Deputy Director with the Smart Systems Institute (SSI) and co-directed the Centre of Social Media Innovations for Communities (CoSMIC) at NUS. He has co-authored a book, seven patents, and more than 300 conference publications, journal articles, and book chapters. His research interests include streaming media architectures, distributed systems, mobile and geo-referenced video management, applications of machine/deep learning, and spatial data management. He is an associate editor for IEEE MultiMedia, ACM Transactions on Multimedia Computing, Communications, and Applications (TOMM), Springer Multimedia Tools and Applications (MTAP), and IEEE Open Journal of the Communications Society (OJ-COMS). He is a distinguished member of the ACM and a senior member of the IEEE. Further information can be found at \url{http://www.comp.nus.edu.sg/~rogerz/}.
\end{IEEEbiography}

\vfill

\end{document}